\documentclass[10pt,twocolumn,letterpaper]{article}

\usepackage[pagenumbers]{iccv} % To force page numbers, e.g. for an arXiv version
\usepackage{soul}
\usepackage{colortbl}
\usepackage{pifont}% http://ctan.org/pkg/pifont
\newcommand{\cmark}{\ding{51}}%
\usepackage[accsupp]{axessibility}
\usepackage{amsmath}

\definecolor{iccvblue}{rgb}{0.21,0.49,0.74}
\usepackage[pagebackref,breaklinks,colorlinks,allcolors=iccvblue]{hyperref}

\definecolor{green}{RGB}{56, 118, 29}
\definecolor{red}{RGB}{204, 0, 0}
\definecolor{burntorange}{RGB}{204, 85, 0}      % Burnt Orange
\definecolor{crimson}{RGB}{220, 20, 60}        % Crimson
\definecolor{teal}{RGB}{0, 128, 128}           % Teal
\definecolor{indigo}{RGB}{75, 0, 130}          % Indigo
\definecolor{royalblue}{RGB}{65, 105, 225}     % Royal Blue
\definecolor{magenta}{RGB}{255, 0, 255}        % Magenta

\usepackage{amsmath}
\usepackage{amssymb}
\usepackage{algorithm}
\usepackage{algpseudocode}
\usepackage{graphicx}
\usepackage{booktabs}
\usepackage{subcaption}
\usepackage{caption}

\usepackage{multirow}
\usepackage{listings}
\usepackage{xcolor}
\usepackage{hyperref}
\hypersetup{urlcolor=red}

\def\paperID{13526} % *** Enter the Paper ID here
\def\confName{ICCV}
\def\confYear{2025}

\title{Supercharged One-step Text-to-Image Diffusion Models with Negative Prompts}
\author{
Viet Nguyen$^{1\boldsymbol{\star}}$, Anh Nguyen$^{1\boldsymbol{\star}}$, Trung Dao$^{1}$, Khoi Nguyen$^{1}$, Cuong Pham$^{1,2}$, Toan Tran$^{1}$, Anh Tran$^{1}$ \\\\ 
$^{1}$Qualcomm AI Research$^{\dagger}$ \quad 
$^{2}$Posts \& Telecom. Inst. of Tech. \\
\vspace{-1.5cm}
}
\def\1n{\mathbf{1}_n}
\def\0{\mathbf{0}}
\def\1{\mathbf{1}}

\def\y{{\bf y}}

\newcommand{\cm}[1]{}

\newcommand{\myheading}[1]{\vspace{1ex}\noindent \textbf{#1}}

\newif\ifshowsolution
\showsolutiontrue
\ifshowsolution

\else

\fi

\begin{document}
\twocolumn[{%
 \renewcommand\twocolumn[1][]{#1}%
 \maketitle
 % \vspace{-6mm}
 \vspace{6mm}
 \centering
 \includegraphics[width=\textwidth]{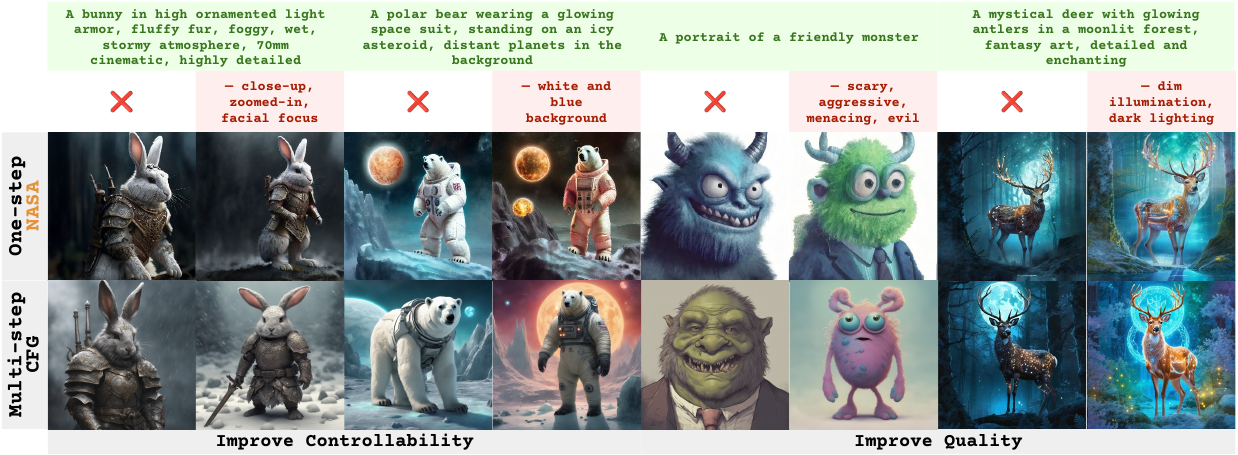}
 \vspace{-4mm}
 \captionof{figure}{
    Negative prompting are a key technique for improving controllability and image quality in text-to-image diffusion models. However, classifier-free guidance (CFG) \cite{ho2022classifier} only supports negative prompt integration in multi-step models due to its iterative nature. Our method, \textbf{NASA}, is \textit{the first} to enable negative prompting in one-step models, allowing for the suppression of unwanted features and providing greater control over image attributes.
   }
    \label{fig:teaser}
   \vspace{2mm}
}]

    \maketitle
    
    \newcommand\blfootnote[1]{%
  \begingroup
  \renewcommand\thefootnote{}\footnote{#1}%
  \addtocounter{footnote}{-1}%
  \endgroup
}

\makeatletter
\def\blfootnote{\gdef\@thefnmark{}\@footnotetext}
\makeatother
    
 \blfootnote{%
  \hspace{-1.7em}$^{\boldsymbol{\star}}$ Equal Contribution \\
  $\dagger$ Qualcomm AI Research is an initiative of Qualcomm Technologies, Inc.%
}

\begin{abstract}The escalating demand for real-time image synthesis has driven significant advancements in one-step diffusion models, which inherently offer expedited generation speeds compared to traditional multi-step methods. However, this enhanced efficiency is frequently accompanied by a compromise in the controllability of image attributes. While negative prompting, typically implemented via classifier-free guidance (CFG), has proven effective for fine-grained control in multi-step models, its application to one-step generators remains largely unaddressed. Due to the lack of iterative refinement, as in multi-step diffusion, directly applying CFG to one-step generation leads to blending artifacts and diminished output quality. To fill this gap, we introduce \textbf{N}egative-\textbf{A}way \textbf{S}teer \textbf{A}ttention (NASA), an efficient method that integrates negative prompts into one-step diffusion models. NASA operates within the intermediate representation space by leveraging cross-attention mechanisms to suppress undesired visual attributes. This strategy avoids the blending artifacts inherent in output-space guidance and achieves high efficiency, incurring only a minimal 1.89\% increase in FLOPs compared to the computational doubling of CFG. Furthermore, NASA can be seamlessly integrated into existing timestep distillation frameworks, enhancing the student's output quality. Experimental results demonstrate that NASA substantially improves controllability and output quality, achieving an HPSv2 score of \textbf{31.21}, setting a new state-of-the-art benchmark for one-step diffusion models. 
% See our project page at: \href{https://snoopi-onestep.github.io/}{\url{snoopi-onestep.github.io/}}
\end{abstract}    
\section{Introduction}
\label{sec:intro}
% Gao2024LuminaT2XTT,Yang2024CogVideoXTD
Diffusion models have recently become popular due to their capacity to produce high-quality, diverse outputs in image \cite{podell2023sdxl,saharia2022photorealistic,balaji2022eDiff-I,Chen2023PixArtFT,autoedit}, 3D \cite{poole2022dreamfusion,wang2023prolificdreamer,lin2023magic3d}, audio \cite{Huang2023MakeAnAudioTG, Evans2024FastTL} and video \cite{xing2023dynamicrafter, BarTal2024LumiereAS} synthesis. Unlike other generative models, such as Generative Adversarial Networks (GANs) \cite{goodfellow2014generative,kang2023gigagan}, diffusion models gradually refine their output through a series of steps, allowing them to achieve diverse yet impressively detailed and realistic outputs. However, this iterative refinement is both time-consuming and computationally demanding, which limits their practicality in real-world applications. As a result, there is an increasing interest in techniques to speed up diffusion models while maintaining output quality. Recent distillation methods \cite{nguyen2024swiftbrush, dao2024swiftbrushv2, yin2024onestep, yin2024improved,dao2025self} compress the entire multi-step generation process into a single step, a breakthrough that enables real-time image synthesis. This speedup makes diffusion models practical for applications like content creation and interactive media, where their slow performance was previously a major drawback. Despite these impressive speed advancements in one-step generators, we have identified a significant limitation that affects all these models: \textbf{the inability to support negative prompting} \cite{blog2023woolf, ban2024understanding}. This feature, which allows users to explicitly exclude unwanted elements, is widely available in multi-step diffusion models but remains \textit{notably absent} in their single-step alternatives.

The most widely adopted approach for negative prompting, classifier-free guidance (CFG) \cite{ho2022classifier} works by generating two parallel noise predictions during each denoising step -- one incorporating the positive prompt and another using the negative prompt. The system then subtracts the negative prediction from the positive one, effectively steering generation away from unwanted attributes. This process relies on the iterative nature of multi-step diffusion, where corrections can be applied and refined over dozens of steps.

One-step generators, however, collapse this entire process into a single forward pass, eliminating the iterative refinement cycles that make CFG possible. Without these multiple correction opportunities, there is no straightforward way to incorporate negative guidance into the generation pipeline. Consequently, users face an unappealing choice between the instant generation that might include unwanted elements or waiting longer for more precisely controlled results. This speed-versus-control dilemma represents the most significant practical barrier to widespread adoption of one-step generators in professional creative workflows, highlighting the urgent need for novel approaches that can bridge this gap.

% To address this issue, a novel approach is needed to incorporate the negative prompt signal directly within the one-step generation process, enabling these models to effectively avoid unwanted features without iterative sampling. 

To address this issue, we introduce a novel method called \textbf{Negative-Away Steer Attention (NASA)} used in inference, which incorporates negative prompt guidance directly into one-step diffusion models. NASA works by adjusting cross-attention layers, effectively reducing unwanted features in generated images. By steering attention within intermediate representation space, NASA offers image control capabilities that were previously achievable only in multi-step diffusion models. 

With our newly developed capability, we have explored integrating negative prompting directly into one-step distillation frameworks, focusing on the SwiftBrush (SB) family of models \cite{nguyen2024swiftbrush, dao2024swiftbrushv2}. The SB approach is compelling for its unique, image-free distillation methodology. By matching teacher model probability distributions and applying sophisticated training techniques, SBv2 can outperform its own teacher on standard image synthesis benchmarks. Despite these technical achievements, SB models still fall short in human preference evaluations. The core challenge lies in the distillation process: without real image supervision, the system lacks clear signals about which generation patterns should be avoided. This is precisely where negative prompting and our NASA framework come in.

Committed to maintaining SB's valuable image-free distillation approach while incorporating guidance sampling to align with human preference, we provide the critical missing component, a way to steer the model away from undesirable outputs during training. NASA enables something previously impossible: guidance sampling within a one-step generator. Through this integration, our model achieves an HPSv2 \cite{Wu2023HumanPS} score of \textbf{31.21}, establishing a new state-of-the-art on human preference metrics for one-step models.

In summary, our contributions are twofold:

\begin{itemize}

    \item We introduce Negative-Away Steer Attention (NASA), the \textit{first method} to integrate negative prompt guidance into \textit{one-step} and \textit{few-step} diffusion models in both \textit{inference} and \textit{distillation} settings.
    
    \item By integrating NASA into the SwiftBrush (SB) distillation framework, we achieve a new state-of-the-art for one-step diffusion models on human preference metrics. This marks the first successful integration of negative prompt guidance directly into one-step model training.
    
\end{itemize}
\section{Related Work}
\label{sec:related_work}
\subsection{Distribution Matching Distillation}
Recent work \cite{nguyen2024swiftbrush, dao2024swiftbrushv2, yin2024onestep, yin2024improved} has explored text-to-image diffusion distillation through distribution matching techniques. These methods directly align the output distributions of teacher and student models, particularly through Variational Score Distillation (VSD). These techniques can be broadly categorized into two approaches: image-dependent training and image-free training.

Image-dependent methods rely on real images for adversarial loss \cite{yin2024improved, sauer2023adversarial} or synthetic ones generated by the teacher models for reconstruction loss \cite{yin2024onestep}. Despite their effectiveness in preserving image quality, these approaches face significant practical limitations: they require substantial storage capacity, consume high memory during training, demand extensive computational resources, and risk mode collapse due to intensive image supervision.

In contrast, image-free approaches \cite{nguyen2024swiftbrush, dao2024swiftbrushv2} have demonstrated promising results by relying solely on prompt text for training, significantly reducing resource requirements while maintaining competitive performance. However, image-free techniques often encounter training instability, particularly within VSD-based frameworks. Existing solutions to stabilize training \cite{yin2024onestep} introduce significant memory and computational overhead, limiting their practical application. Our work addresses this gap by introducing a simple yet effective method that enhances training stability without additional computational burden, making image-free variational score distillation produce better models. 

\subsection{Sampling Guidance with Negative Prompts}
Classifier-free guidance (CFG) \cite{ho2022classifier} and negative prompting \cite{blog2023woolf, ban2024understanding} are key techniques for content control during denoising. In multi-step diffusion models, these methods enhance desired image features and suppress undesired ones, improving visual fidelity and reducing artifacts \cite{dhariwal2021diffusion, katzir2023noisefree, Armandpour2023ReimagineTN}. Score-based approaches like NFSD \cite{katzir2023noisefree} employ negative prompting to steer generated samples toward real-image distributions, while SDS-Bridge \cite{McAllister2024RethinkingSD} uses negative prompts to model source distributions more accurately. Despite their success, negative prompting has yet to be integrated into one-step or few-step diffusion models. To bridge this gap, we introduce NASA, the \textit{first method} to enable negative prompting in \textit{one-step} and \textit{few-step} diffusion models, enhancing control in faster generation settings.

\section{Preliminaries}
\label{sec:background}
\textbf{Diffusion Models}~\cite{ho2020denoising, song2021scorebased, rombach2022high} produce high-quality images by progressively denoising inputs. This technique involves a forward process that gradually adds noise to the data and a reverse process that reconstructs the original data by removing noise. While the original diffusion models act on the image space \cite{ho2020denoising, song2021scorebased}, LDM \cite{rombach2022high} processes on the latent space produced by a pretrained VAE for more efficient computation. In particular, starting with a data point $\mathbf{x}_0$ sampled from an unknown distribution $q(\mathbf{x}_{0})$, the forward process gradually diffuses $\mathbf{x}_0$ into a standard Gaussian noise $\mathbf{x}_T \sim \mathcal N(\mathbf{0}, \mathbf{I})$  through $T$ consecutive timesteps, where $\mathbf I$ is the identity matrix. At each time-step $t$, a noisier version of $\mathbf{x}_0$, denoted as $\mathbf{x}_t$, is drawn from $q_t(\mathbf{x}_{t}|\mathbf{x}_{0}) = \mathcal{N}(\alpha_t\mathbf{x}_{0}, \sigma^2_{t}\mathbf{I})$ using a standard Gaussian noise $\boldsymbol{\epsilon} \sim \mathcal N(0,\mathbf{I})$:
\begin{equation}
    \mathbf{x}_t = \alpha_t \mathbf{x}_0 + \sigma_t \boldsymbol{\epsilon}, \quad \forall t \in \{0, \dots, T\},
    \label{eq:forward}
\end{equation}
where $\{(\alpha_t, \sigma_t)\}_{t=1}^T$ defines the noise schedule, with boundary conditions $(\alpha_0, \sigma_0) = (1, 0)$ for the clean sample and $(\alpha_T, \sigma_T) = (0, 1)$ for pure noise.

Conversely, the reverse process aims to gradually reconstruct the original data by denoising the input over $T$ steps, starting from an initial random Gaussian noise $\mathbf{x}_T \sim \mathcal N(\mathbf{0}, \mathbf{I})$.
The model is trained by minimizing the difference between the noise estimated by the model, $\boldsymbol{\epsilon}_\phi$, parameterized by $\phi$, and the actual noise in \cref{eq:forward}:
\begin{equation}\label{eq:unconditional_loss}
    \min_{\phi} \mathbb{E}_{t \sim \mathcal{U}(0, T), \boldsymbol{\epsilon} \sim \mathcal{N}(0, I)} \Vert \boldsymbol{\epsilon}_{\phi}(\mathbf{x}_t, t) - \boldsymbol{\epsilon} \Vert_2^2.
\end{equation} 
\noindent \textbf{Text-to-Image Diffusion Models} guide the sampling process using an additional text prompt $\mathbf{y}$ to produce outputs that are both photorealistic and aligned with the provided textual descriptions. The training objective, slightly modified from the unconditional loss in \cref{eq:unconditional_loss}, is defined as:
\begin{equation}\label{eq:conditional_loss}
    \min_{\phi} \mathbb{E}_{t \sim \mathcal{U}(0, T), \boldsymbol{\epsilon} \sim \mathcal{N}(0, I), \mathbf{y}} \Vert \boldsymbol{\epsilon}_{\phi}(\mathbf{x}_t, t, \mathbf{y}) - \boldsymbol{\epsilon} \Vert_2^2.
\end{equation}
\noindent \textbf{Classifier-free Guidance (CFG)} \cite{ho2022classifier} is an inference technique designed to enhance the quality of generated images by blending the predictions from both a conditional and an unconditional model. At each sampling step, CFG adjusts the denoiser’s output using a control parameter $\kappa > 1$, allowing controlled guidance that aligns more closely with the desired conditions:
\begin{equation}
    \hat{\boldsymbol{\epsilon}}_\phi(\mathbf{x}_t, t, \y, \kappa) =  (1 - \kappa) \, \boldsymbol{\epsilon}_\phi(\mathbf{x}_t, t) + \kappa \, \boldsymbol{\epsilon}_\phi(\mathbf{x}_t, t, \y).
\label{eq:cfg_null}
\end{equation}

\noindent \textbf{Negative Prompting} \cite{blog2023woolf, ban2024understanding} provides enhanced control by suppressing unwanted features in the generated content. Instead of producing an unconditional output, the model generates an output conditioned on the negative prompt $\mathbf{y}^{-}$, as follows:
\begin{equation}
    \hat{\boldsymbol{\epsilon}}_\phi(\mathbf{x}_t, t, \mathbf{y}, \mathbf{y}^{-}, \kappa) =  (1 - \kappa) \, \boldsymbol{\epsilon}_\phi(\mathbf{x}_t, t, \mathbf{y}^{-}) + \kappa \, \boldsymbol{\epsilon}_\phi(\mathbf{x}_t, t, \mathbf{y}).
\label{eq:cfg_neg}
\end{equation}

\noindent \textbf{Variational Score Distillation (VSD)} is a powerful framework that utilizes pretrained diffusion-based text-to-image models to enhance text-based generation across both 3D~\cite{wang2023prolificdreamer} and 2D~\cite{nguyen2024swiftbrush, dao2024swiftbrushv2,yin2024onestep, yin2024improved} domains. 
The central aim of VSD is to align the renderings of a differentiable generator with the probability density of plausible images as guided by the 2D diffusion model. To accomplish this, VSD employs a two-teacher approach that uses a fixed pretrained diffusion model $\boldsymbol{\epsilon}_\psi$ and an adaptive LoRA-based teacher $\boldsymbol{\epsilon}_\pi$. While training, the student model $\boldsymbol{f}_\theta$ produces 2D images $\hat{\mathbf{x}}_0 = \boldsymbol{f}_\theta(\mathbf{z}, \mathbf{y})$ using an input noise $\mathbf{z} \sim \mathcal N(\mathbf{0}, \mathbf{I})$ and a text prompt $\mathbf{y}$. Noisy images $\hat{\mathbf{x}}_t = \alpha_t \hat{\mathbf{x}}_0 + \sigma_t \boldsymbol{\epsilon}$ are then fed into both teacher models. The LoRA teacher model $\boldsymbol{\epsilon}_\pi$ aligns with the student distribution by minimizing a denoising L2 loss on single-step samples. This arrangement supports robust and adaptive guidance suitable for a variety of generative architectures. The gradient of the learning objective with respect to the student's parameters $\theta$ is defined as:
\begin{equation}
\begin{split}
    \nabla_{\theta} \mathcal{L}_{\text{VSD}} = \mathbb{E}_{t, \mathbf{y}, \mathbf{z},\kappa=C} & \bigg[ \, w(t) \left(\hat{\boldsymbol{\epsilon}}_{\psi}(\hat{\mathbf{x}}_t, t, \mathbf{y}, \kappa) \right. \\
    & \left. - \hat{\boldsymbol{\epsilon}}_{\pi}(\hat{\mathbf{x}}_t, t, \mathbf{y}, \kappa \right) \frac{\partial \boldsymbol{f}_{\theta}(\mathbf{z}, \mathbf{y})}{\partial \theta} \bigg],
\end{split}
\label{eq:vsd}
\end{equation}
where $C$ is a constant and $w(t)$ is the time-dependent weight that adjusts the contribution of each timestep, aligning the student’s outputs with the diffusion model’s predictions.
Through alternating updates of the student and LoRA teacher, VSD enables efficient, high-fidelity generation across both text-to-image and text-to-3D tasks.

\begin{table}[t]
\centering
\small
\begin{tabular}{cccccc}
\toprule
\textbf{NP} & \textbf{Anime} & \textbf{Photo} & \textbf{CA} & \textbf{Paintings} & \textbf{Average}\\
\midrule
 & 29.62 & 29.17 & 28.79 & 28.69 & 29.07 \\
\rowcolor{cyan!4}\cmark & 31.66 & 29.17 & 30.58 & 30.38 & 30.42 \\
\bottomrule
\end{tabular}
\vspace{-1mm}
\caption{The effect of negative prompts (NP) on the HPSv2 score of PixArt-$\alpha$ \cite{chen2024pixartalpha}. Using generic negative prompts to remove visual artifacts (e.g., ``worst quality", ``blurry", ``bad anatomy") leads to a significant improvement. Higher scores indicate better alignment with human preferences. 
}
\label{tab:hps_teacher}
\vspace{-2mm}
\end{table}

\section{Observations}
\subsection{Motivation}
\label{sec:observations}
\begin{figure}[t]
\centering
\includegraphics[width=\linewidth]{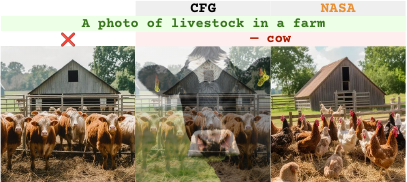} 
\vspace{-5mm}
\caption{The inherent single-pass nature of one-step diffusion models renders them incompatible with direct CFG, resulting in artifact-laden outputs.} 
\label{fig:naiveCFG} 
\vspace{-5mm}
\end{figure}

\textbf{Positive prompt is not enough.} While vision generative models like diffusion models excel at incorporating positive attributes, they fundamentally struggle with understanding negation. Consider requesting ``A portrait of a friendly monster, not scary" (\cref{fig:teaser}), the model typically fails to exclude the specified negative attribute ``scary". Negative prompting addresses this limitation by creating a separate, dedicated input for exclusions -- completely removing negation terms from the positive prompt and processing them independently. This clear separation allows the model to accurately distinguish what to avoid generating, providing significantly better control over the final output.

\noindent \textbf{Community Need.} The growing use of negative prompting in the generative AI community further highlights its importance. Platforms like Civit AI \cite{civitai} and various open-source repositories \cite{comfyui,sdwebui, forge} have shown that using both types of prompts together results in more refined, higher-quality model outputs that align more closely with user specifications. Users can clearly specify both desired and unwanted attributes in a prompt pair, which is particularly useful when working with models that generate complex or highly detailed content. Experimentally, \Cref{tab:hps_teacher} and \Cref{fig:teaser} confirm that using negative prompts effectively enhances image quality of the model.

\noindent \textbf{Critical Gap.} Despite the established need and widespread adoption of negative prompts in multi-step diffusion models and the rapidly growing prominence of one-step approaches, a significant limitation persists: \textit{no existing one-step methods} effectively support negative prompting. Our proposed method fills this gap by providing a simple, efficient, and training-free mechanism that integrates negative prompt into one-step diffusion models to (1) improve \textit{controllability} and (2) enhance the \textit{quality} of the output image.

\label{sec:nasa}
\begin{figure*}[t] 
\centering 
\includegraphics[width=.97\linewidth]{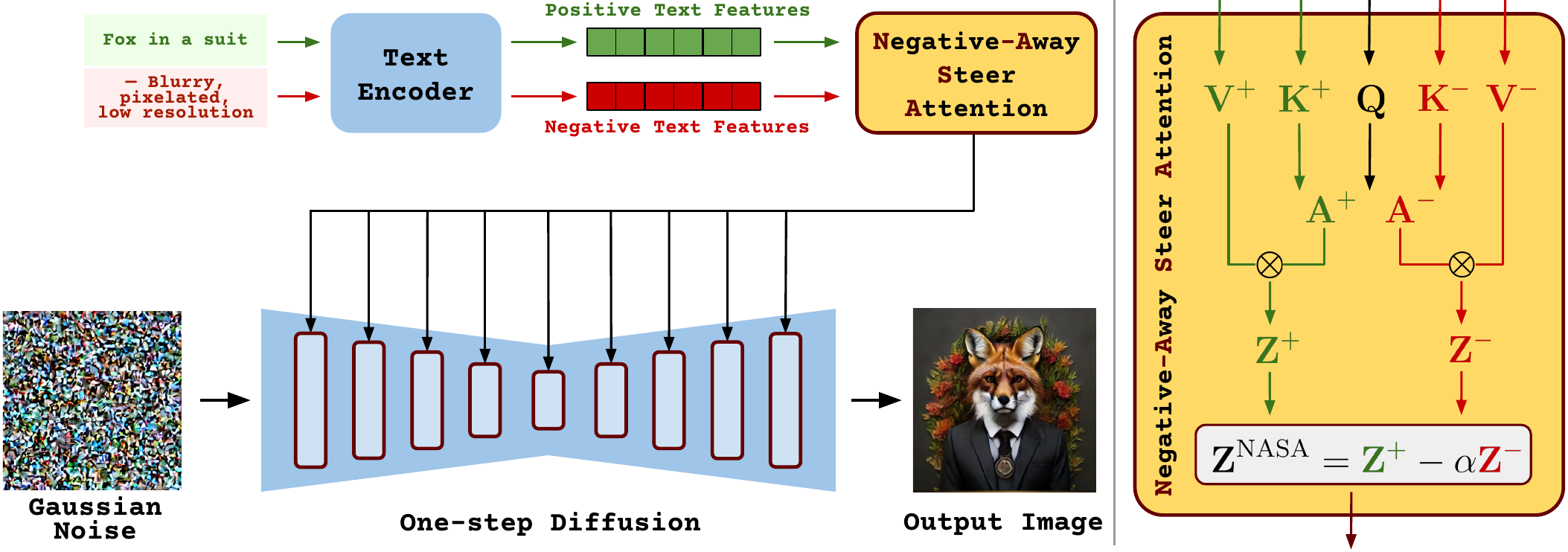} 
\caption{\textit{\textbf{Left:}} An overview of the Negative-Away Steer Attention (NASA) pipeline. Positive (\textcolor{green}{green}) and negative (\textcolor{red}{red}) prompts are fed into a text encoder to generate positive and negative text features. The NASA module then processes these features, which adjusts the one-step diffusion model to steer the output image away from the negative features, refining it based on the positive features. \textit{\textbf{Right:}} The details of the NASA module. It processes queries ($\mathbf{Q}_l$) in layer $l$, note we will omit the subscript $l$ in subsequent notations to improve readability,  with positive ($\mathbf{V}^+, \mathbf{K}^+$) and negative ($\mathbf{V}^-, \mathbf{K}^-$) key-value pairs to create positive ($\mathbf{Z}^+$) and negative ($\mathbf{Z}^-$) attention outputs. The final output ($\mathbf{Z}^{\text{NASA}}$) is calculated by subtracting the weighted negative features ($\mathbf{Z}^{-}$) from the positive features ($\mathbf{Z}^{+}$).} 
\vspace{-4mm}
\label{fig:pipe_nasa} 
% \vspace{-4mm}
\end{figure*}

\subsection{Challenges}
A widely used method for implementing negative prompting is the CFG mechanism. However, applying CFG directly to one-step diffusion models presents significant challenges. CFG was designed for multi-step models, where the output is iteratively refined, allowing for gradual adjustments using both negative and unconditional outputs. In contrast, one-step models generate images in a single pass without the iterative denoising process, making them inherently unsuitable for this method. Direct application of CFG to these models leads to undesirable image blending, resulting in unnatural artifacts, as shown in \cref{fig:naiveCFG}. This highlights a crucial limitation: the CFG mechanism, though effective in multi-step models, is not directly translatable to one-step models without significant modifications.
\section{Proposed Method}
Our proposed methodology is detailed in \cref{subsec:nasa} for the core mechanism. We present two primary implementation strategies for NASA: a training-free approach (NASA\text{--}I) and a distillation-based approach (NASA\text{--}T), which are presented in \cref{subsec:inference} and \cref{subsec:training}, respectively.

\subsection{NASA: \textbf{N}egative-\textbf{A}way \textbf{S}teer \textbf{A}ttention}
\label{subsec:nasa}
In contrast to CFG implementations that operate within the output space (synthesized image), as commonly practiced in multi-step diffusion models, we propose a novel approach that strategically relocates the guidance mechanism to the \textit{intermediate representation space} of the diffusion model. 

Specifically, cross-attention layers in these models capture semantic connections between image and text features~\cite{hertz2022prompt}, making them ideal for controlling the feature alignment. We hypothesize that cross-attention layer features are particularly well-suited for this approach. We selectively manipulate these cross-attention maps to attenuate attention responses associated with negative prompts relative to those elicited by positive prompts, thereby filtering out unwanted semantic attributes.

This \textit{representation-space} guidance strategy offers a crucial advantage: it enables the exclusion of undesirable elements \textit{prior} to the final image synthesis stage. This removal mechanism significantly mitigates the risk of generating blending artifacts, which are common byproducts of image-space manipulation techniques.

Our\textit{ representation-space} guidance also extends to architectures like FLUX \cite{flux2024}, which replace traditional cross-attention with joint self-attention blocks. \cite{wei2024enhancing} show that these blocks still embed distinct image-text cross-attention operations. This allows NASA to be seamlessly applied, proving its robustness across different backbones.

The overall pipeline of the NASA mechanism is illustrated in \cref{fig:pipe_nasa}. Given the latent feature representation at a given layer $l$, denoted as $\mathbf{Z}_l$, and positive text features $\mathbf{c}_p$ extracted from CLIP text encoder, we define the query, key, and value matrices as $\mathbf{Q}_l = \mathbf{Z}_l \mathbf{W}_{q,l}$, $\mathbf{K}_{l}^+ = \mathbf{c}_p \mathbf{W}_{k,l}$, and $\mathbf{V}_{l}^+ = \mathbf{c}_p \mathbf{W}_{v,l}$, where $\mathbf{W}_{q,l}$, $\mathbf{W}_{k,l}$, and $\mathbf{W}_{v,l}$ are projection matrices at current layer $l$. Analogously, for negative text features $\mathbf{c}_n$, we use the same projection matrices to define the corresponding key and value matrices for negative prompt conditioning: $\mathbf{K}_{l}^- = \mathbf{c}_n \mathbf{W}_{k,l}$ and $\mathbf{V}_{l}^- = \mathbf{c}_n \mathbf{W}_{v,l}$.
\begin{table}[t]
\centering
\small
\setlength{\tabcolsep}{5pt}
  \begin{tabular}{c cc cc cc}
    \toprule
    \multirow{2}{*}{Method} & \multicolumn{2}{c}{\textbf{FLUX.1-schnell}} & \multicolumn{2}{c}{\textbf{SDXL-LCM}} & \multicolumn{2}{c}{\textbf{SDXL-DMD2}} \\
    \cmidrule(lr){2-3} \cmidrule(lr){4-5} \cmidrule(lr){6-7} 
      & 4 steps & 1 step & 4 steps & 1 step & 4 steps & 1 step \\
    \midrule
    None & 23\% & 44\% & 43\% & - & 27\% & 25\% \\
    CFG & 30\% & 0\% & 14\% & - & 25\% & 0\% \\    
    \rowcolor{cyan!4} \textbf{NASA\text{--}I} & 100\% & 99\%  & 97\% & - & 100\% & 100\% \\
    \bottomrule
  \end{tabular}
  \vspace{-2mm}
  \caption{Comparison of success rate of unwanted feature removal in generated images. Note that SDXL-LCM \cite{luo2023latent} does not support one-step inference.}
  \vspace{-6mm}
  \label{tab:neg_prompt_acc}
\end{table} 
\noindent The NASA mechanism is formulated as follows:
\begin{align} 
\mathbf{Z}_l^{+} &= 
\text{Softmax}\left(\frac{\mathbf{Q}_l(\mathbf{K}_{l}^{+})^{\top}}{\sqrt{d}}\right)\mathbf{V}_{l}^{+}, \\ 
\mathbf{Z}_l^{-} &= 
\text{Softmax}\left(\frac{\mathbf{Q}_l(\mathbf{K}_{l}^{-})^{\top}}{\sqrt{d}}\right)\mathbf{V}_{l}^{-},\\
&\mathbf{Z}^{\text{NASA}}_{l} = \mathbf{Z}_l^{+} - \alpha\cdot\mathbf{Z}_l^{-},
\end{align}
where $\alpha$ is the scale factor to control the degree of negative feature removal. 

\subsection{NASA\text{--}I: Training-free Setting}
\label{subsec:inference}
\noindent \textbf{Applicable to One-/Few-Step Models.} To assess the effectiveness of NASA\text{--}I, we carry out experiments with pretrained one-step and few-step models, such as LCM \cite{luo2023latent}, DMD2 \cite{yin2024improved} and FLUX.1-schnell \cite{flux2024}. We design a small set of positive and negative prompts and generate 100 images per model, both with and without negative prompts. We then calculate the percentage of images that successfully exclude the feature specified by the negative prompt. For example, given the positive prompt ``A photo of a person'' and the negative prompt ``male'', we measure the percentage of generated images that do not depict a male person. The average results across all prompt sets are summarized in \cref{tab:neg_prompt_acc}. Our findings show that while traditional CFG is ineffective for one-step and few-step models, NASA\text{--}I provides precise control over unwanted features in generated images. Additionally, this capability is also qualitatively demonstrated in \cref{fig:teaser}, \cref{fig:scaleNASA_main} and \cref{fig:qualitative_flux}. 
% (further details are provided in the Appendix)

\noindent \textbf{Efficiency Considerations.} As described in \cref{subsec:nasa}, we reuse the projection matrices of the positive prompt for the negative prompt, ensuring that no additional model parameters are introduced and keeping the model size unchanged. Furthermore, applying NASA-I to all one-step and few-step models introduces only a minor computational overhead, increasing FLOPs by just 1.89 \%. This represents a significant efficiency improvement over CFG, which doubles the computational cost due to the need for separate forward passes for positive and negative prompts. Consequently, NASA\text{--}I enables negative prompt guidance in one-step diffusion models without introducing substantial latency, making it a practical solution for real-time applications.

\subsection{NASA\text{--}T: Distillation Setting}
\label{subsec:training}
Given their effectiveness in improving image quality during inference, their potential utility in model distillation is worth exploring. In this section, we introduce a method that integrates the negative prompt into one-step diffusion distillation. While our approach is applicable to various distillation methods, we select SwiftBrush (SB) \cite{nguyen2024swiftbrush,dao2024swiftbrushv2}, an image-free method, as the baseline for improvement due to its efficiency. As described in \cref{eq:vsd}, SB updates the student model using the gradient of the VSD loss. 
To incorporate the negative prompt $\mathbf{y}^{-}$, we modify the existing formulation by replacing $\hat{\boldsymbol{\epsilon}}_{\psi}(\hat{\mathbf{x}}_t, t, \mathbf{y}, \kappa)$ (see \cref{eq:cfg_null}) with $\hat{\boldsymbol{\epsilon}}_{\psi}(\hat{\mathbf{x}}_t, t, \mathbf{y}, \mathbf{y}^{-}, \kappa)$ (see \cref{eq:cfg_neg}).
In this way, the teacher model provides explicit guidance to the student model, helping it learn to steer away from undesirable features. Moreover, we incorporate the NASA mechanism into the student model, enabling it to process negative prompts during training. This allows the student model to better recognize and suppress unnatural artifacts, leading to improved learning and enhanced image quality. Note that with the LoRA teacher model $\hat{\boldsymbol{\epsilon}}_{\pi}$, we can choose whether to incorporate negative prompts. Empirically, we found that not using them yields better results. Detailed results and discussions are provided in \cref{subsec:ablation}.
\vspace{-2mm}
\section{Experiments}
\label{sec:exp}

\begin{table*}
\centering
\resizebox{.8\textwidth}{!}{
\begin{tabular}{l|cc|ccccl}
\toprule
\textbf{Dataset} & \multicolumn{2}{c|}{\textbf{NegOpt}} & \multicolumn{5}{c}{\textbf{HPSv2}} \\
\midrule
\textbf{Method} & \textbf{CLIP${}^{+}$} $\uparrow$ & \textbf{CLIP${}^{-}$} $\downarrow$ & \textbf{Anime} $\uparrow$ & \textbf{Photo} $\uparrow$ & \textbf{CA} $\uparrow$ & \textbf{Paintings} $\uparrow$ & \textbf{Average} $\uparrow$\\
\midrule
\rowcolor{gray!20} \multicolumn{8}{c}{PixArt-$\alpha$-based backbone} \\
PixArt-$\alpha$ \cite{chen2024pixartalpha} [Teacher] & 0.35 & 0.05 & 29.62 & 29.17 & 28.79 & 28.69 & 29.07 \\
\rowcolor{cyan!4} YOSO \cite{luo2024sample} & 0.36 & 0.08 & 28.75 & 28.06 & 28.52 & 28.57 & 28.48 \\
\rowcolor{cyan!4} + \textbf{NASA\text{--}I} & 0.36 & 0.06 & 28.74 & 28.05 & 28.56 & 28.60 & 28.49 \textcolor{blue}{(+0.01)} \\
DMD \cite{yin2024onestep} & 0.35 & 0.08 & 29.31 & 28.67 & 28.46 & 28.41 & 28.71 \\
+ $\text{CFG}=1.5$ & 0.34 & 0.09 & 30.02 & 27.07 & 28.36 & 28.07 & 28.38 \textcolor{red}{(-0.33)} \\
+ $\text{CFG}=2.5$ & 0.31 & 0.13 & 26.74 & 23.86 & 25.13 & 24.66 & 25.10 \textcolor{red}{(-3.61)} \\
+ \textbf{NASA\text{--}I} & 0.35 & 0.05 & 29.33 & 28.71 & 28.49 & 28.53 & 28.77 \textcolor{blue}{(+0.06)} \\
\rowcolor{cyan!4} SBv2${}^{*}$ & 0.36 & 0.09 & 32.19 & 29.09 & 30.39 & 29.69 & 30.34 \\
\rowcolor{cyan!4} + \textbf{NASA\text{--}I} & 0.36 & 0.06 & 32.60 & 29.58 & 31.09 & 30.65 & 30.98 \textcolor{blue}{(+0.64)} \\
\rowcolor{cyan!4} + \textbf{NASA\text{--}T} & 0.35 & 0.08 & 32.33 & 29.26 & 30.75 & 30.10 & 30.61 \textcolor{blue}{(+0.27)} \\
\rowcolor{cyan!4} + \textbf{NASA\text{--}T} + $\text{CFG}=1.5$ & 0.34 & 0.10 & 29.47 & 26.50 & 28.22 & 27.68 & 27.97 \textcolor{red}{(-2.37)} \\
\rowcolor{cyan!4} + \textbf{NASA\text{--}T} + \textbf{NASA\text{--}I} & 0.35 & 0.06 & \textbf{32.66} & \textbf{29.67} & \underline{31.31} & \underline{30.71} & \underline{31.09} \textcolor{blue}{(+0.75)} \\
\rowcolor{cyan!4} + \textbf{NASA\text{--}T} + \textbf{NASA\text{--}I} ($\alpha=1$) & 0.35 & 0.05 & \underline{32.65} & \underline{29.65} & \textbf{31.45} & \textbf{31.06} & \textbf{31.21} \textcolor{blue}{(+0.87)} \\
\midrule 
\rowcolor{gray!20} \multicolumn{8}{c}{Stable Diffusion 1.5-based backbone} \\
SDv1.5 \cite{rombach2022high} [Teacher] & 0.31 & 0.06 & 26.51 & 27.19 & 26.06 & 26.12 & 26.47 \\
\rowcolor{cyan!4} InstaFlow-0.9B \cite{liu2023instaflow} & 0.33 & 0.11 & 26.10 & 26.62 & 25.92 & 25.95 & 26.15 \\
\rowcolor{cyan!4} + \textbf{NASA\text{--}I} & 0.33 & 0.10 & 26.15 & 26.68 & 25.93 & 25.98 & 26.19 \textcolor{blue}{(+0.04)}\\
DMD2 \cite{yin2024improved} & 0.31 & 0.10 & 26.39 & 27.00 & 25.80 & 25.82 & 26.25 \\
+ \textbf{NASA\text{--}I} & 0.31 & 0.09 & 26.39 & 27.02 & 25.80 & 25.83 & 26.26 \textcolor{blue}{(+0.01)} \\
\rowcolor{cyan!4} SBv2${}^{*}$ & 0.32 & 0.09 & 27.18 & 27.58 & 26.69 & 26.62 &  27.02 \\
\rowcolor{cyan!4} + \textbf{NASA\text{--}I} & 0.32 & 0.08 & 27.19 & 27.59 & 26.71 & 26.63 & 27.03 \textcolor{blue}{(+0.01)} \\
\rowcolor{cyan!4} + \textbf{NASA\text{--}T} & 0.35 & 0.10 & \underline{27.64} & \underline{27.94} & \textbf{27.19} & \underline{27.03} & \underline{27.45} \textcolor{blue}{(+0.43)} \\
\rowcolor{cyan!4} + \textbf{NASA\text{--}T} + \textbf{NASA\text{--}I} & 0.35 & 0.09 & \textbf{27.65} & \textbf{27.97} & \underline{27.18} & \textbf{27.04} & \textbf{27.46} \textcolor{blue}{(+0.44)} \\
\midrule
\rowcolor{gray!20} \multicolumn{8}{c}{Stable Diffusion 2.1-based backbone} \\
SDv2.1 \cite{rombach2022high} [Teacher] & 0.33 & 0.06 & 27.48 & 26.89 & 26.86 & \underline{27.46} & 27.17 \\
 SBv2 \cite{dao2024swiftbrushv2} & 0.34 & 0.10 & 27.25 & 27.62 & 26.86 & 26.77 & 27.13 \\
 + \textbf{NASA\text{--}I} & 0.35 & 0.08 & 27.44 & 27.84 & 26.91 & 27.02 & 27.30 \textcolor{blue}{(+0.17)} \\
\rowcolor{cyan!4} SBv2${}^{*}$ & 0.37 & 0.10 & 27.56 & 27.84 & 26.97 & 27.03 & 27.35 \\
\rowcolor{cyan!4} + \textbf{NASA\text{--}I} & 0.37 & 0.08 & 27.70 & 28.00 & 27.12 & 27.22 & 27.51 \textcolor{blue}{(+0.16)} \\
\rowcolor{cyan!4} + \textbf{NASA\text{--}T} & 0.36 & 0.09 & \underline{28.00} & \underline{28.65} & \underline{27.44} & 27.26 & \underline{27.84} \textcolor{blue}{(+0.49)} \\
\rowcolor{cyan!4} + \textbf{NASA\text{--}T} + \textbf{NASA\text{--}I} & 0.36 & 0.07 & \textbf{28.04} & \textbf{28.68} & \textbf{27.51} & \textbf{27.50} & \textbf{27.93} \textcolor{blue}{(+0.58)} \\
\bottomrule
\end{tabular}
 }
\caption{Quantitative comparisons between our method and previous works. \textbf{NASA\text{--}I} refers to our training-free method introduced in \cref{subsec:inference}, while \textbf{NASA\text{--}T} denotes our distillation-based approach described in \cref{subsec:training}. SBv2${}^{*}$ refers to our reimplementation of SBv2 with a randomly sampled CFG scale $\kappa$.}
\label{tab:all_score}
\end{table*}

\begin{figure*} 
\centering 
\includegraphics[width=.85\linewidth]{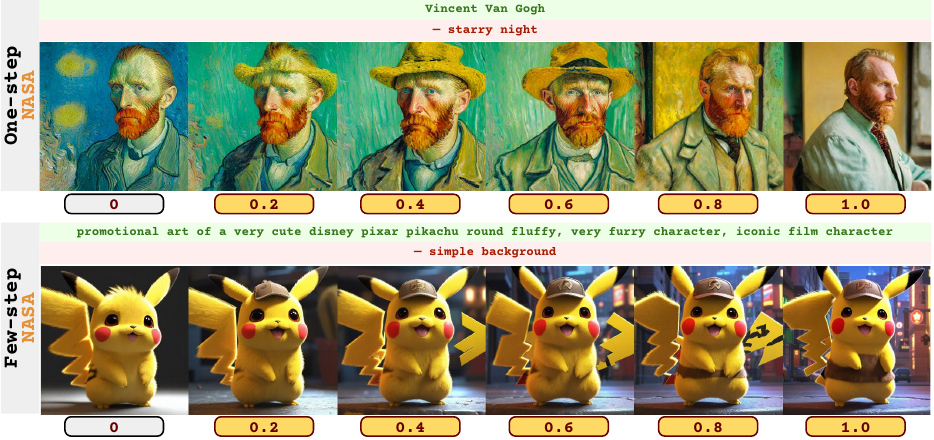} 
\caption{Effect of different scale values (0.0 to 1.0) in NASA-I with SDXL-DMD2 \cite{yin2024improved} for both one-step and few-step settings, illustrating the progressive influence on visual details and composition.} 
\vspace{-2mm}
\label{fig:scaleNASA_main} 
\end{figure*}

\begin{figure*}[t]
  \centering
  \includegraphics[width=\textwidth]{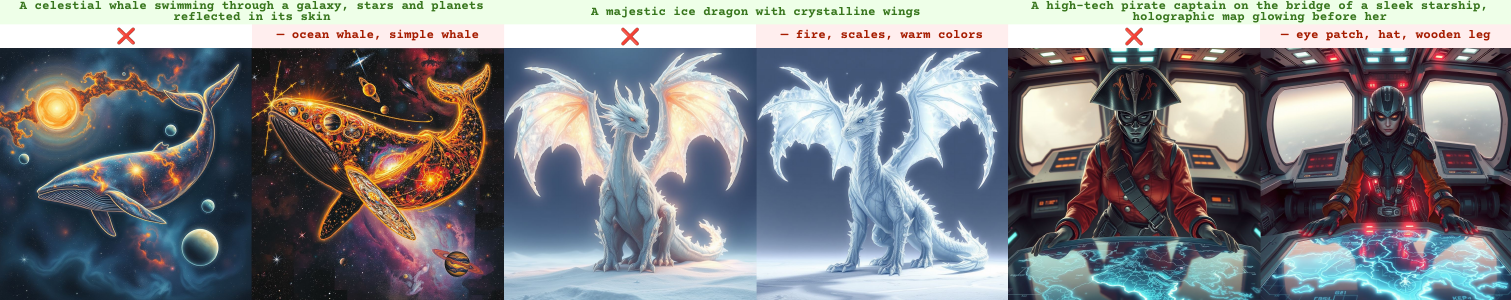}  
  \vspace{-5mm}
  \caption{Qualitative results of NASA-I in FLUX.1-schnell \cite{flux2024}.}
  \label{fig:qualitative_flux}
  \vspace{-2mm}
  % \vspace{-10pt}
\end{figure*}

\begin{figure*}[t]
  \centering
  % --- Figure on the left in a minipage ---
  \begin{minipage}[b]{0.73\textwidth}
      \centering
      \includegraphics[width=\textwidth]{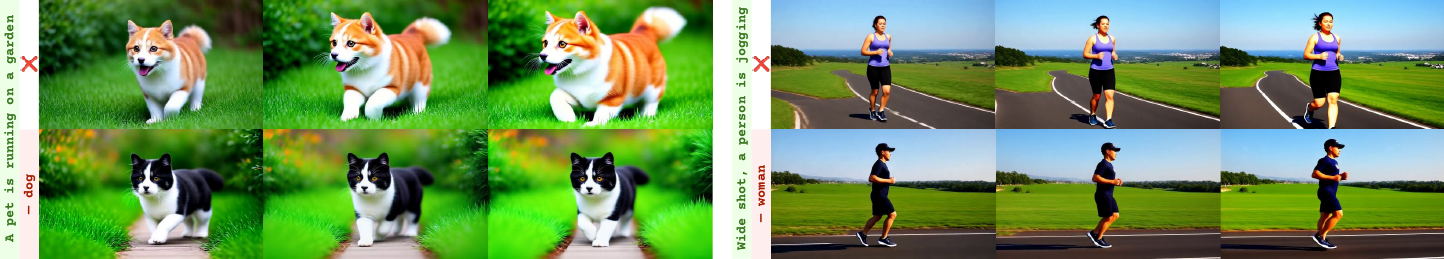}
      % \vspace{1mm}
      \captionof{figure}{Qualitative results of NASA-I for text-to-video generation in CausVid \cite{yin2025causvid}.}
      \label{fig:qualitative_video}
  \end{minipage}
  \vspace{-1mm}
  \hfill
  % --- Table on the right in a minipage ---
  \begin{minipage}[b]{0.25\textwidth}
      \centering
      \scriptsize
      \setlength{\tabcolsep}{2pt}
      \begin{tabular}{l cc}
        \toprule
        \multirow{2}{*}{Method} & \multicolumn{2}{c}{VBench-Long benchmark} \\
        \cmidrule(lr){2-3}
        & \parbox{1.5cm}{\centering Aesthetic Quality $\uparrow$} & \parbox{1.5cm}{\centering Imaging Quality $\uparrow$} \\
        \midrule
        None & 61.98 & 67.12 \\
        % CFG & - & - \\    
        \rowcolor{cyan!4}\textbf{NASA--I} & 63.33 & 67.36 \\
        \bottomrule
      \end{tabular}
      % \vspace{-1mm}
      \captionof{table}{CausVid quality metrics.}
      \label{tab:causvid_quality}
  \end{minipage}
  \vspace{-1mm}
\end{figure*}

\subsection{Experimental Setup}
\myheading{Metrics and Benchmark.}
In diffusion model evaluation, Fréchet Inception Distance (FID) \cite{Heusel2017GANsTB} on the zero-shot MS-COCO2014 \cite{mscoco} benchmark has been the standard metric. However, recent studies \cite{ren2024hypersd, chen2024pixart, yin2024improved} suggest that FID in this benchmark does not always correlate with actual image quality and adopt human-focused alignment metrics like Human Preference Score v2 (HPSv2) \cite{Wu2023HumanPS} for benchmarking. Our experiments also follow this conclusion, so we choose HPSv2 as the primary metric for evaluating the models. Moreover, due to the absence of negative prompts in the MS-COCO2014 benchmark, we utilize the NegOpt dataset from \cite{ogezi2024optimizing}, which contains pairs of positive and negative prompts. From this dataset, we randomly sample 30K prompt pairs to construct a subset for evaluation. Text alignment is then assessed using CLIP scores, where CLIP${}^{+}$ evaluates alignment with the positive prompt and CLIP${}^{-}$ evaluates alignment with the negative prompt.

\myheading{Training Datasets.} Following SBv2 \cite{dao2024swiftbrushv2}, we use total of 3.3M prompts from JourneyDB \cite{sun2023journeydb} and a subset of LAION \cite{schuhmann2022laion}\footnote{All datasets were downloaded and evaluated at MovianAI.}. For negative prompts in the distillation setting (\cref{subsec:training}), we construct a dataset of 10K negative prompts derived from commonly used terms like ``bad photo, duplicate, low resolution, low quality''.

\myheading{Implementation Details.}
We base our method on SBv2 \cite{dao2024swiftbrushv2} with our proposed modifications in \cref{subsec:training}. We utilize three backbones SDv1.5 \cite{rombach2022high}, SDv2.1 \cite{rombach2022high} and PixArt-$\alpha$ \cite{chen2024pixartalpha} as the frozen teachers and LoRA teachers are initialized with rank $r = 64$ and scaling $\gamma = 128$. The learning rates are set to $1e^{-6}$ for the student model and $1e^{-3}$ for the LoRA teacher, using the AdamW optimizer \cite{loshchilov2018decoupled}. All our training is conducted on four NVIDIA A100 40GB GPUs with the total batch size of 64 for the SD models and 32 for PixArt-$\alpha$. Moreover, \cite{chadebec2024flash} found that instead of using a fixed CFG scale $\kappa$ during training, sampling $\kappa$ from a uniform distribution within a given range improves training robustness. Based on this insight, we set the teacher models’ CFG scales as follows: $\kappa \sim \mathcal{U}(0.5, 4)$ for SDv1.5 and SDv2.1, and $\kappa \sim \mathcal{U}(0.5, 3)$ for PixArt-$\alpha$. For NASA\text{--}T, in each iteration, we uniformly sample the scale $\alpha \sim \mathcal{U}(0, 1)$ to help the student model adapt to different conditions and enhance generalization. For NASA\text{--}I, unless explicitly specified, we set $\alpha = 0.1$ for SDv1.5-based models, $\alpha = 0.2$ for SDv2.1-based models, and $\alpha = 0.5$ for PixArt-$\alpha$-based models. Further details can be found in the Appendix.
\vspace{-1mm}
\begin{table}[h]
\centering
\small
\setlength{\tabcolsep}{6pt} % Adjust spacing between columns for better fit
\begin{tabular}{l| ccccc}
\toprule
\textbf{NASA} & & & & \cmark & \cmark \\
\textbf{Teacher} & & \cmark & \cmark & \cmark & \cmark \\
\textbf{LoRA Teacher} & & & \cmark & & \cmark \\
\midrule
\textbf{HPSv2 Avg. $\uparrow$} & 30.34 & 29.84 & 29.80 & \textbf{30.61} & 29.89 \\
\bottomrule
\end{tabular}
\vspace{-1mm}
\caption{Ablation studies on the integration of negative prompts in model distillation using the PixArt-$\alpha$ backbone.}
\label{tab:ablation_neg}
\vspace{-2mm}
\end{table}

\begin{figure} 
\centering
\includegraphics[width=.95\linewidth]{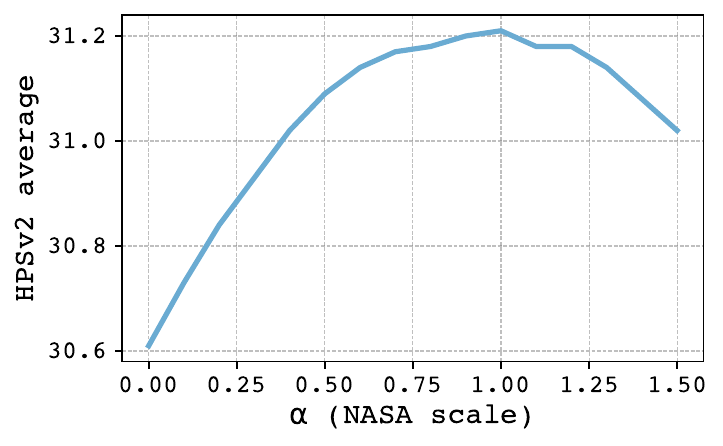} 
\vspace{-3mm}
\caption{Ablation study on the NASA scale using our distilled model based on the PixArt-$\alpha$ backbone.} 
\vspace{-5mm}
\label{fig:ablation_alpha} 
\end{figure}

\subsection{Main Results}
\label{subsec:quantitative}
\vspace{-1mm}
\myheading{Quantitative Results.} \Cref{tab:all_score} presents a comparison between our approach and prior distilled text-to-image diffusion models. Using our reimplementation of SBv2 \cite{dao2024swiftbrushv2} as the baseline, NASA\text{--}T consistently enhances image quality across all tested backbones. Compared to other methods, our model achieves the highest HPSv2 scores on average, demonstrating its clear superiority in image generation quality. This suggests that integrating complex loss functions like DMD2 \cite{yin2024improved} may not always be necessary for achieving state-of-the-art results. Moreover, while applying CFG directly degrades one-step model outputs, NASA\text{--}I further enhances image quality across all models when using negative prompts like ``worst quality, low quality, ugly, duplicate, out of frame, deformed, blurry, bad anatomy, watermark''. Notably, our method on the PixArt-$\alpha$-based backbone achieves a record-breaking HPSv2 score of \textbf{31.21}, surpassing all other one-step diffusion models baselines. Regarding the NegOpt dataset, with the support of NASA\text{--}I, all models demonstrate a reduction in the CLIP${}^{-}$ score while maintaining the CLIP${}^{+}$ score. This indicates our method effectively eliminates unwanted features in generated outputs while preserving desired traits. 

\myheading{Qualitative Results.} As discussed in \cref{subsec:inference},  NASA\text{--}I is a versatile, training-free method for both one-step and few-step diffusion models. \Cref{fig:scaleNASA_main} and \Cref{fig:qualitative_flux} present qualitative results of integrating NASA\text{--}I into the SDXL-DMD2 \cite{yin2024improved} and FLUX.1-schnell \cite{flux2024} models, respectively. When using a conditional negative prompt, the generated images retain key semantic features while effectively suppressing unwanted attributes specified by the negative prompt. 

\myheading{Application to Text-to-Video Models.} To demonstrate the versatility of our approach, we extended NASA-I to a few-step video diffusion model, CausVid \cite{yin2025causvid}. As shown in \cref{fig:qualitative_video}, our method effectively suppresses unwanted visual attributes while preserving key semantic content and temporal consistency. Quantitative evaluation on the VBench-Long \cite{huang2023vbench} benchmark further validates this. \Cref{tab:causvid_quality} shows that using NASA\text{--}I with common negative prompts like ``low quality, bad anatomy, blurry, distortion, ugly, low resolution, unclear details'' improves both the aesthetic and imaging quality scores of the generated videos. This highlights NASA\text{--}I's potential as a general-purpose tool for enhancing guided generation across different modalities.

\subsection{Ablation Studies}
\label{subsec:ablation}
\Cref{tab:ablation_neg} presents an ablation on integrating negative prompts during model distillation with the PixArt-$\alpha$ \cite{chen2024pixartalpha} backbone. The baseline is our reimplementation of SBv2 with a randomly sampled CFG scale $\kappa$. Simply applying negative prompts only in the teacher and/or the LoRA teacher models slightly degrades the HPSv2 score, as the student model is not explicitly trained on this negative conditioning. In contrast, incorporating NASA with negative prompts during training allows the student model to better capture and utilize this conditioning, leading to improved performance. Furthermore, we find that integrating negative prompts solely into the teacher model yields better results than applying them to both the teacher and LoRA teacher models.

Using our PixArt-$\alpha$-based distilled model as the base, we varied the NASA scale $\alpha$ and evaluated the results with HPSv2 on the benchmark dataset. \Cref{fig:ablation_alpha} demonstrates that NASA\text{--}I generally enhances image synthesis, resulting in outputs that better align with human preferences. However, if the NASA scale $\alpha$ is set too high, the negative attributes may overly dominate the positive attributes, adversely affecting the final image quality.
\section{Discussion and Conclusion}
\label{sec:conclusion}
This paper introduces NASA, the first method to integrate negative prompt guidance into one-step and few-step diffusion models for both inference and distillation. By leveraging cross-attention adjustments, NASA effectively suppresses unwanted features in generated images. Experimental results show that NASA enhances image quality and outperforms strong baselines, establishing a new state-of-the-art in one-step model distillation based on human preference metrics. While NASA's effectiveness relies on selecting an appropriate scale for suppressing negative features, this parameter does not require extensive tuning.
{
    \small
    \bibliographystyle{ieeenat_fullname}
    \bibliography{main}
}
\clearpage
\maketitlesupplementary

\section{Implementation Details}
\myheading{NASA.}
We provide the PyTorch pseudo-code for the NASA algorithm, as outlined in Section 5.1 in the main paper. The implementation is straightforward and is provided in \cref{alg:nasa_attention}.
\begin{algorithm*}[!ht]
\small
\caption{\small NASA.}
\label{alg:nasa_attention}
\definecolor{codeblue}{rgb}{0.25,0.5,0.5}
\definecolor{codekw}{rgb}{0.85, 0.18, 0.50}
\lstset{
  backgroundcolor=\color{white},
  basicstyle=\fontsize{7.5pt}{7.5pt}\ttfamily\selectfont,
  columns=fullflexible,
  breaklines=true,
  captionpos=b,
  commentstyle=\fontsize{7pt}{7pt}\bfseries\color{codeblue},
  keywordstyle=\fontsize{7pt}{7pt}\bfseries\color{codekw},
  escapechar={|}, 
  % numbers=left,
  % stepnumber=1,    
  % firstnumber=1,
  % numberfirstline=true,
}
\begin{lstlisting}[language=Python]
class NASA_AttnProcessor(nn.Module):
    def __init__(self, nasa_scale=1.0):
        super().__init__()
        self.nasa_scale = nasa_scale

    def __call__(
        self, attn, z, emb, neg_emb, attn_mask, neg_attn_mask, temb, *args, **kwargs,
    ):
        # Input preparation
        residual = z
        if attn.spatial_norm is not None:
            z = attn.spatial_norm(z, temb)
        input_ndim = z.ndim
        if input_ndim == 4:
            bsz, channel, height, width = z.shape
            z = z.view(bsz, channel, height * width).transpose(1, 2)
            
        bsz, sequence_length, _ = z.shape if emb is None else emb.shape
        _, neg_sequence_length, _ = z.shape if neg_emb is None else neg_emb.shape
        
        if attn_mask is not None:
            attn_mask = attn.prepare_attention_mask(attn_mask, sequence_length, bsz)
            attn_mask = attn_mask.view(bsz, attn.heads, -1, attn_mask.shape[-1])
        if neg_attn_mask is not None:
            neg_attn_mask = attn.prepare_attention_mask(neg_attn_mask, neg_sequence_length, bsz)
            neg_attn_mask = neg_attn_mask.view(bsz, attn.heads, -1, neg_attn_mask.shape[-1])
            
        if attn.group_norm is not None:
            z = attn.group_norm(z.transpose(1, 2)).transpose(1, 2)
            
        query = attn.to_q(z)
        if emb is None:
            emb = z
        else:
            if attn.norm_cross:
                emb = attn.norm_emb(emb)
                neg_emb = attn.norm_emb(neg_emb)
                
        # Compute cross-attention for positive embedding
        key = attn.to_k(emb)
        value = attn.to_v(emb)
        inner_dim = key.shape[-1]
        head_dim = inner_dim // attn.heads
        query = query.view(bsz, -1, attn.heads, head_dim).transpose(1, 2)
        key = key.view(bsz, -1, attn.heads, head_dim).transpose(1, 2)
        value = value.view(bsz, -1, attn.heads, head_dim).transpose(1, 2)
        if attn.norm_q is not None:
            query = attn.norm_q(query)
        if attn.norm_k is not None:
            key = attn.norm_k(key)

        z = F.scaled_dot_product_attention(
            query, key, value, attn_mask=attn_mask, dropout_p=0.0, is_causal=False
        )
        z = z.transpose(1, 2).reshape(bsz, -1, attn.heads * head_dim)
        z = z.to(query.dtype)
        
        # Compute cross-attention for negative embedding
        neg_key = attn.to_k(neg_emb)
        neg_value = attn.to_v(neg_emb)
        neg_key = neg_key.view(bsz, -1, attn.heads, head_dim).transpose(1, 2)
        neg_value = neg_value.view(bsz, -1, attn.heads, head_dim).transpose(1, 2)

        neg_z = F.scaled_dot_product_attention(
            query, neg_key, neg_value, attn_mask=neg_attn_mask, dropout_p=0.0, is_causal=False
        )
        neg_z = neg_z.transpose(1, 2).reshape(bsz, -1, attn.heads * head_dim)
        neg_z = neg_z.to(query.dtype)

        # Equation 10 in main paper
        z_nasa = z - self.nasa_scale * neg_z
        z_nasa = attn.to_out[0](z_nasa)
        z_nasa = attn.to_out[1](z_nasa)
        
        if input_ndim == 4:
            z_nasa= z_nasa.transpose(-1, -2).reshape(bsz, channel, height, width)
        if attn.residual_connection:
            z_nasa = z_nasa + residual
        z_nasa = z_nasa / attn.rescale_output_factor

        return z_nasa
\end{lstlisting}
\end{algorithm*}

\myheading{NASA\text{--}T.}
Following SBv2 \cite{dao2024swiftbrushv2}, we also use the clamped CLIP loss with a margin of $\tau = 0.37$, starting with a weight of 0.1 and gradually reducing to zero. \Cref{tab:hyperparams} provides the additional hyperparameters for training SBv2 with NASA\text{--}T.

\begin{table}[h]
\centering
\small
\setlength{\tabcolsep}{4pt}
\resizebox{\linewidth}{!}{
\begin{tabular}{cccc}
\toprule 
\textbf{Hyperparameter} & \textbf{SDv1.5} & \textbf{SDv2.1} & \textbf{PixArt-$\alpha$} \\
\midrule 
Dataset & JDB + LAION & JDB + LAION & JDB \\
Batch size & 64 & 64 & 32  \\
Training iterations & 60k & 40k & 50k  \\
Mixed-Precision (BF16) & Yes & Yes & Yes  \\
$\kappa$ & $\mathcal{U}(0.5, 4)$ & $\mathcal{U}(0.5, 4)$ &$\mathcal{U}(0.5, 3)$  \\
$\alpha$ & $\mathcal{U}(0, 1)$ & $\mathcal{U}(0, 1)$ & $\mathcal{U}(0, 1)$ \\
Clip weight & 0.1 & 0.1 & 0.1 \\
$\tau$ & 0.37 & 0.37 & 0.37 \\
$lr$ of student & $1e^{-6}$ & $1e^{-6}$ & $1e^{-6}$ \\
$lr$ of LoRA teacher & $1e^{-3}$ & $1e^{-3}$ & $1e^{-3}$ \\
LoRA rank $r$ & 64 & 64 & 64 \\
LoRA scaling $\gamma$ & 128 & 128 & 128 \\
\bottomrule
\end{tabular}
}
\caption{Hyperparameters used for training SBv2 \cite{dao2024swiftbrushv2} with NASA\text{--}T.}
\label{tab:hyperparams}
\end{table}

\section{More Qualitative Results}
\begin{figure*}
\includegraphics[width=\linewidth]{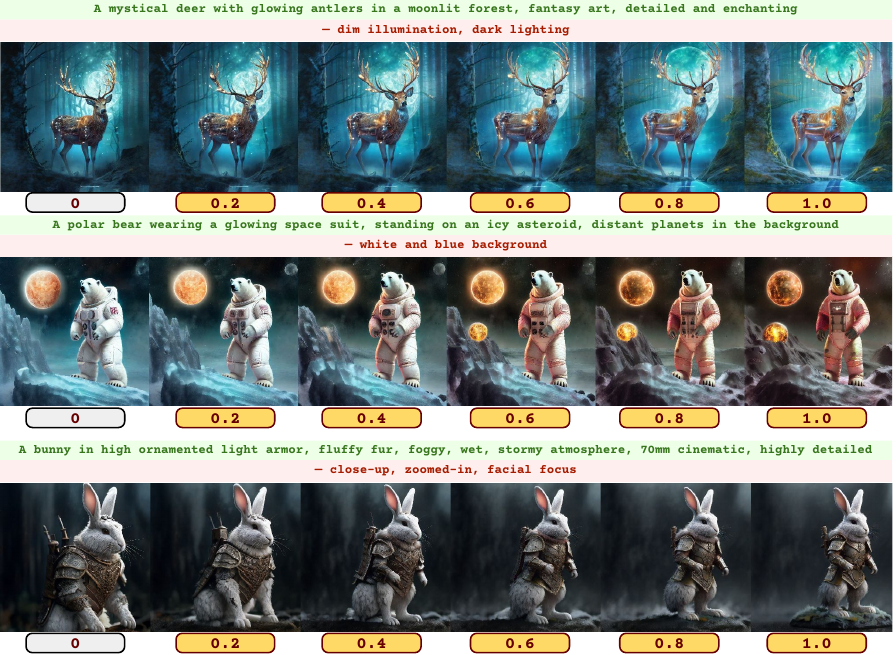} 
\caption{Additional qualitative images of applying NASA\text{--}I to SDXL-DMD2 1-step.} 
\label{fig:supp_1step} 
\end{figure*}

\begin{figure*}
\includegraphics[width=\linewidth]{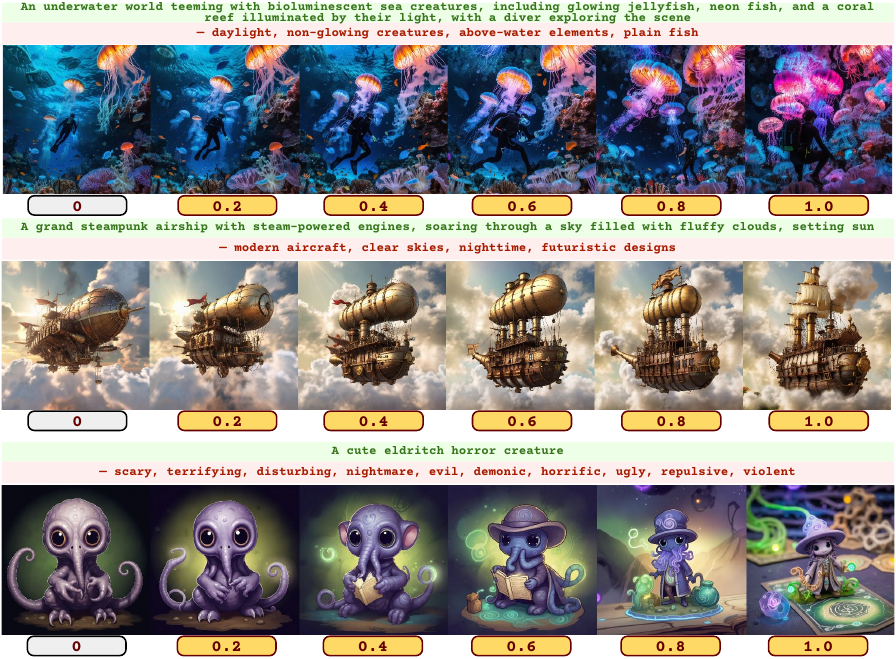} 
\caption{Additional qualitative images of applying NASA\text{--}I to SDXL-DMD2 4-step.} 
\label{fig:supp_4step} 
\end{figure*}

\Cref{fig:supp_1step} and \Cref{fig:supp_4step} illustrate the visual effect of different scale values $\alpha$ ranging from 0.0 to 1.0 in the NASA method when being applied on SDXL-DMD2. A higher scale value $\alpha$ results in more effective removal of the feature specified by the negative prompt.

\end{document}